\documentclass[runningheads]{llncs}
\usepackage[T1]{fontenc}
\usepackage{graphicx}
\usepackage{algorithm}
\usepackage{algorithmic}
\usepackage{breqn}
\usepackage{newfloat}
\usepackage{listings}
\usepackage{multirow}
\usepackage{tabularray}
\usepackage{color,soul}
\usepackage{comment}
\usepackage{xcolor}
\begin{document}
\title{Balancing the Scales: Enhancing Fairness in Facial Expression Recognition with Latent Alignment\thanks{Supported by Kwikpic AI Solutions.}}
\titlerunning{Balancing the Scales: Enhancing Fairness in FER}
%
\newcommand{\repeatthanks}{\textsuperscript{\thefootnote}}

\author{Syed Sameen Ahmad Rizvi\thanks{denotes equal contribution and joint first authorship}\orcidID{0000-0002-3919-5074} \and
Aryan Seth\repeatthanks\orcidID{0009-0005-4171-4901} \and
Pratik Narang\orcidID{0000-0003-1865-3512}}
\authorrunning{S.S.A. Rizvi and A. Seth et. al.}
%
\institute{ Birla Institute of Technology and Science, Pilani\\
\email{\{p20190412,f20212221,pratik.narang\}@pilani.bits-pilani.ac.in}\\
}
\maketitle              
\begin{abstract}
Automatically recognizing emotional intent using facial expression has been a thoroughly investigated topic in the realm of computer vision. Facial Expression Recognition (FER), being a supervised learning task, relies heavily on substantially large data exemplifying various socio-cultural demographic attributes.  Over the past decade, several real-world in-the-wild FER datasets that have been proposed were collected through crowd-sourcing or web-scraping. However, most of these practically used datasets employ a manual annotation methodology for labelling emotional intent, which inherently propagates individual demographic biases. Moreover, these datasets also lack an equitable representation of various socio-cultural demographic groups, thereby inducing a class imbalance. Bias analysis and its mitigation have been investigated across multiple domains and problem settings; however, in the FER domain, this is a relatively lesser explored area. This work leverages representation learning based on latent spaces to mitigate bias in facial expression recognition systems, thereby enhancing a deep learning model's fairness and overall accuracy.

\end{abstract}
\section{Introduction}
Facial expression recognition (FER) has been an extensively explored problem in the field of deep learning and computer vision. In the past decade, numerous proposed FER datasets have made it easier to approach facial expression recognition as a supervised deep-learning task. Deep learning requires large and diverse datasets for efficaciously modelling data distribution. However, such a supervised learning strategy necessitates substantial training data that reflects a wide range of socio-cultural demographic characteristics.

Over the past decade, various real-world, in-the-wild datasets have been proposed using web-scraped/crowd-sourced images. A crucial drawback of employing such a data-driven method for expression recognition lies in its susceptibility to biases present in the datasets, particularly those that disproportionately affect specific demographic groups.\cite{cite1,cite2}. Facial Expression Recognition requires human annotations per image, which propagates annotative biases and prejudices.  Moreover, most real-world in-the-wild datasets lack proportionate representation of different demographic attributes such as race, age, and gender. Another crucial factor contributing to bias in FER datasets is crowd-sourced annotation. Each annotator possesses their own bias with respect to understanding facial expressions in varied demographics. However, given the enormous size of datasets, these biases are often assumed to be components of random noise.\cite{cite3,cite4}.

In practice, however, people may harbour systematic and demographic biases, especially when inadequately trained with proper demographic and psychological knowledge; they may incorporate such biases into their annotations \cite{r4}. Bias is defined as systematic mistakes that result in unjust outcomes during a decision-making process. In the realm of deep learning, this can originate from multiple factors, such as data collection methodology, algorithm design, and biased human annotation \cite{correction1}. A deep learning model trained on such datasets would inherently propagate bias, thus making it unfair. Fairness in the context of deep learning refers to the absence of bias or discrimination in deep learning systems; however, achieving it can be difficult since deploying a real-world deep learning solution can propogate biases that can emerge in such systems.

Annotative biases combined with class and demographic imbalances increase bias and reduce equal-odds fairness for attributes such as gender, ethnicity, etc. Therefore, examining the biases within datasets and designing algorithms to mitigate them becomes crucial.
Considering age as a protected attribute in datasets, we observe that adolescents are represented positively (such as happy) \cite{r4}; on the contrary, senior citizens are represented more negatively (such as sad and disgusted). This causes models to be biased, with adolescents being classified more frequently to positive expressions, viz-a-viz, and senior citizens being predicted to negative expressions.

Bias analysis and its mitigation strategies have gained good traction among researchers working in the facial analysis domain. However, in the FER domain, this is a relatively less explored area \cite{correction2,r1}. This work is our attempt to tackle and mitigate this bias, therefore increasing fairness in a deep learning model.
 The major contributions of this work include:
\begin{itemize}
\item A novel latent alignment technique with an architecture that generates improved latent representations, mitigates bias, and improves accuracy for FER.
\item A novel training technique and loss function that uses Variational Autoencoders and an adversarial discriminator with perceptual loss for bias mitigation and a CNN backbone for expression classification.
\item Conducting extensive evaluation on two commonly used datasets (RAF-DB \cite{r3} and CelebA\cite{celeba}) and multiple protected attributes in both separate and combined techniques, mitigating bias towards gender, race, and age, setting new state-of-the-art results and competitive performance.
\end {itemize}  
This paper is an extended version of our Student Abstract published at AAAI-24\cite{aaai}, which, to the best of our knowledge, is the first attempt to explore representation learning using latent spaces in mitigating biases in the facial expression domain. This paper provides more comprehensive experimentation with an additional dataset (CelebA\cite{celeba}), detailed results on the interplay between different protected attributes, and better insights into our methodology and training approach. \\
The rest of the paper is organised as follows: Section \ref{sec:related_work} discusses some recent notable works in bias mitigation. Section \ref{sec:meth} describes the methodology adopted, including the training methodology, loss functions, and classification model employed. Section \ref{sec:exp} presents our experimental results, the evaluation metric and analysis of datasets. Section \ref{ablation} provides a component-wise ablation study of our proposed architecture. Section \ref{sec:conc} concludes the work and presents directions for future work.

\section{Recent Works}
\label{sec:related_work}
Bias in Machine learning has attracted wider attention in recent years, with the rapid growth in the deployment of real-world machine learning applications. Extensive surveys\cite{I,II,III,IV} have been done to study bias and its mitigation strategies. In this section, we discuss some of the notable methods for mitigating biases. In literature\cite{III} three types of bias mitigation strategies have been discussed, namely, pre-processing, in-processing, and post-processing methods.

\textit{Pre-processing Methods:} Calmon et al. \cite{V} proposed an optimized pre-processing strategy that modifies the data features and labels. Zemel et al. \cite{VI} proposed a mitigation strategy that learns fair representations by formulating fairness as an optimization problem of finding good representations of the data while obfuscating any information about membership in the protected group. Feldman et al. \cite{VII} proposed disparate impact remover, where feature values were modified while preserving rank ordering to improve overall fairness.

\textit{In-processing:} Kamishima et al. proposed a prejudice remover mechanism \cite{VIII} that leverages a discrimination-aware regularization approach to the learning objective that can be applied to any prediction algorithm with probabilistic discriminative models. Zhang et al. \cite{IX} proposed a strategy that learns fair representations by including a variable for the group of interest and simultaneously learning a predictor and an adversary. Meta Fair Classifier \cite{X} proposes a meta-algorithm for classification that takes fairness constraints as input and returns an optimised classifier.

 \textit{Post-processing:} Reject option Classification \cite{XI} presents a discriminative aware classification, which essentially aims at the prediction that carries a higher degree of uncertainty and thereby assigns favourable outcomes to unprivileged groups and unfavourable outcomes to privileged groups. The strategy of calibrated equalized odds \cite{XII} is designed to optimise the calibrated classifier score outputs. Its goal is to identify probabilities that can be used to alter output labels while maintaining an objective of equalized odds.

Some other techniques to tackle \textit{dataset bias} include transfer learning\cite{XIII}, adversarial mitigation\cite{XIV,XV}, and domain adaptation \cite{XVI,XVII,XVIII}. Various strategies have been proposed to eliminate or prevent models from acquiring misleading or unwanted correlations. A post-hoc correction technique \cite{XIX} that imposes an equality of odds constraint on previously learnt predictor. In the domain of deep learning, two popular techniques are the tweaking of loss functions to impose penalties on unfairness\cite{XX}, and adversarial learning \cite{IX,XXI,XXII}. These techniques aim to learn a fair representation that is devoid of any information related to protected attributes.

\textbf{Bias mitigation in Facial Expression Recognition:} Bias mitigation in facial expression recognition is a relatively less-explored area. With the exponential increase in computing capabilities over the past decade, many datasets and algorithms have been proposed for automatically recognizing facial expressions. However, most of these in the wild real-world datasets are either web-scraped or crowd-sourced. These datasets often have two major limitations \cite{XXIII}. Firstly, most datasets have class imbalances; i.e. people with varied socio-cultural-ethnic identities are inadequately represented among various classes. Secondly, since these large numbers of scraped images are manually labelled by a group of annotators, a personal bias is inherently a part of the dataset. 

Some of the existing works that have tackled bias and it's mitigation in facial expression recognition include a facial Action Unit (AUs) calibrated FER approach \cite{XXIV}, an attribute aware and a disentangled method \cite{r1}. Zeng et al. \cite{XXVI} proposed a Meta-Face2Exp framework that utilized large unlabelled facial recognition datasets.

\section{Methodology}
\label{sec:meth}

We propose a two-part model for mitigating bias. Recognizing that CNNs tend to learn from all input features, for the first part of the model we propose a Variational Autoencoder (VAE) to encode the images into a latent space. The images corresponding to each protected attribute in the dataset will each have a corresponding latent space. Our goal is to minimize the distance between these latent spaces so that each latent encodes only the information relevant to expression classification. 
We propose to utilize a Variational Autoencoder with shared weights for all protected attributes where the inter-latent domain gap is reduced using an adversarial discriminator. We denote the Encoder part as E and the Generator part as G.
We introduce a two-part model to address bias mitigation. Given CNNs' propensity to assimilate all input features, our initial model component employs a Variational Autoencoder (VAE) to encode images belonging to protected attributes into the common latent space. The goal is to minimise disparities between these latent spaces, ensuring they contain information relevant to expression classification.\\
Summarising the methodology:
\begin{itemize}
    \item The main cause of bias is that models tend to learn protected attributes as features.
    \item Our model solves this by generating a latent that has forgotten the protected attribute.
    \item This is done by overlapping the latent spaces of data points belonging to different protected attributes; this overlap is done using the discriminator.
\end{itemize}

\begin{figure*}[t]
\begin{center}
\label{fig:methodology_diag}
\label{fig1}
\includegraphics [scale=0.55]{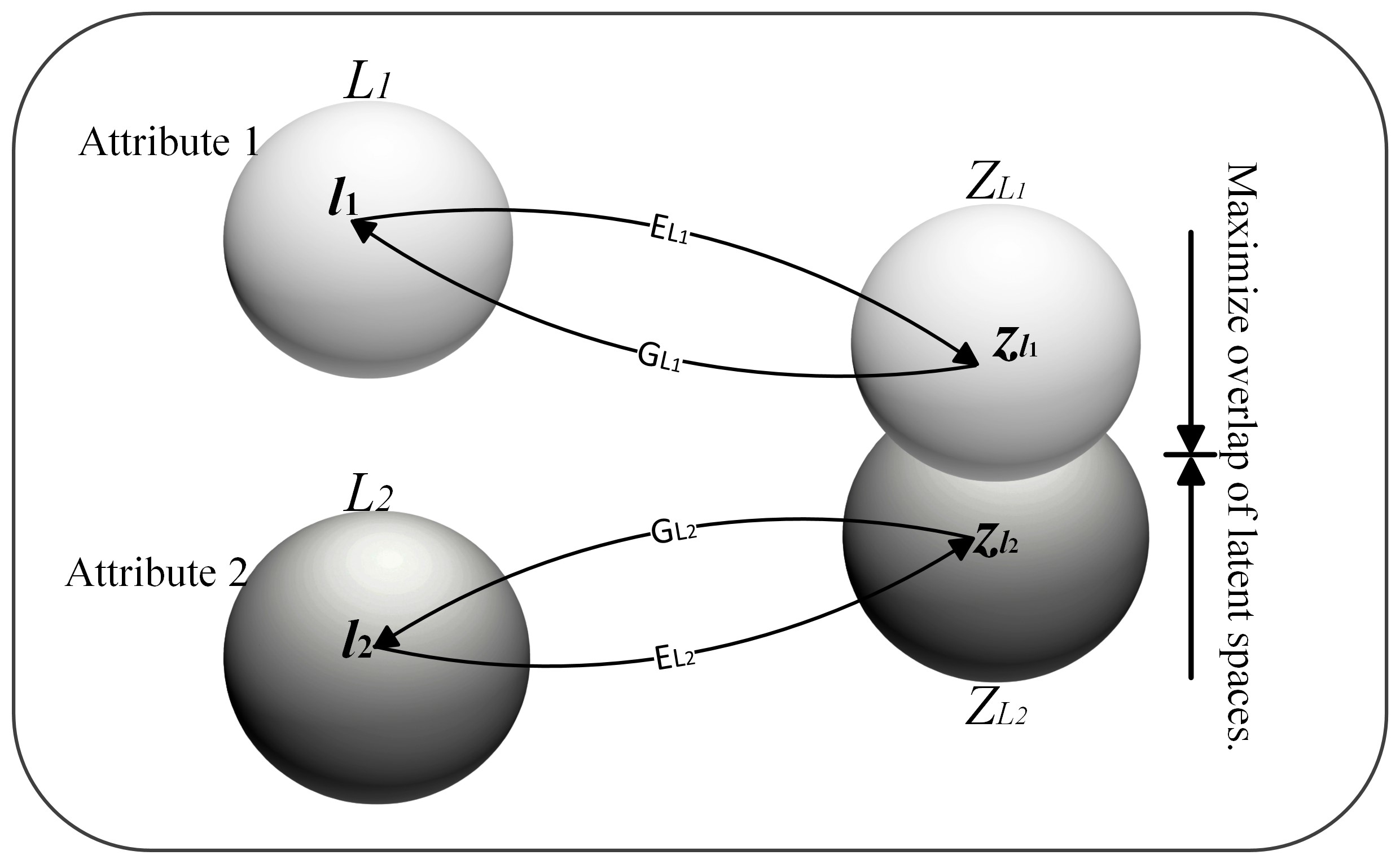}
\centering
\caption{Architecture for Attribute Disentanglement. $L_i$ represents data having the attribute $q_i$. $Z_{L_i}$ is the latent representation of $L_i$. $E_{L_i}$ is a VAE with shared weights $\forall i$. 'E' refers to the Encoder module, which compresses the input image into a latent that does not contain information about the protected attribute. 'G' refers to the Generator, which is a reconstruction module that converts the latent back to the original image.}

\end{center}
\end{figure*} 

\textbf{Attribute Disentanglement -} We propose a shared-weight Variational Autoencoder across all protected attributes, mitigating inter-latent domain disparities through an adversarial discriminator. In this context, we denote the Encoder and Generator components as \lq E' and \lq G'. This is demonstrated in Fig. \ref{fig1}, where $q_i$ is a protected attribute such as gender.
\begin{equation}
\label{VAE_Objective}
\begin{aligned}
\mathcal{L}_{\mathrm{VAE}}(x) =&\mathrm{KL}\left(z_x \mid x\right) \| \mathcal{N}(0, I)) 
+\mathcal{L^{\mathrm{Latent}}_{\mathrm{VAE}, \mathrm{D}}}(x)\\
&+\alpha\left\|G_j^\phi(\hat{y})-G_j^\phi(y)\right\|_F^2
\end{aligned}
\end{equation}
Equation \ref{VAE_Objective} is the objective function for the VAE. The first component consists of KL-divergence that penalizes deviation of the latent distribution from a Gaussian Distribution. The second component is discriminator loss, which measures whether the discriminator can predict the protected attribute class. The final component is Style-Reconstruction Loss \cite{r5}.\\
\subsubsection{Classification Model } 
\begin{figure*}[t]
\begin{center}
\label{fig:methodology_arch}
\includegraphics [scale=0.55]{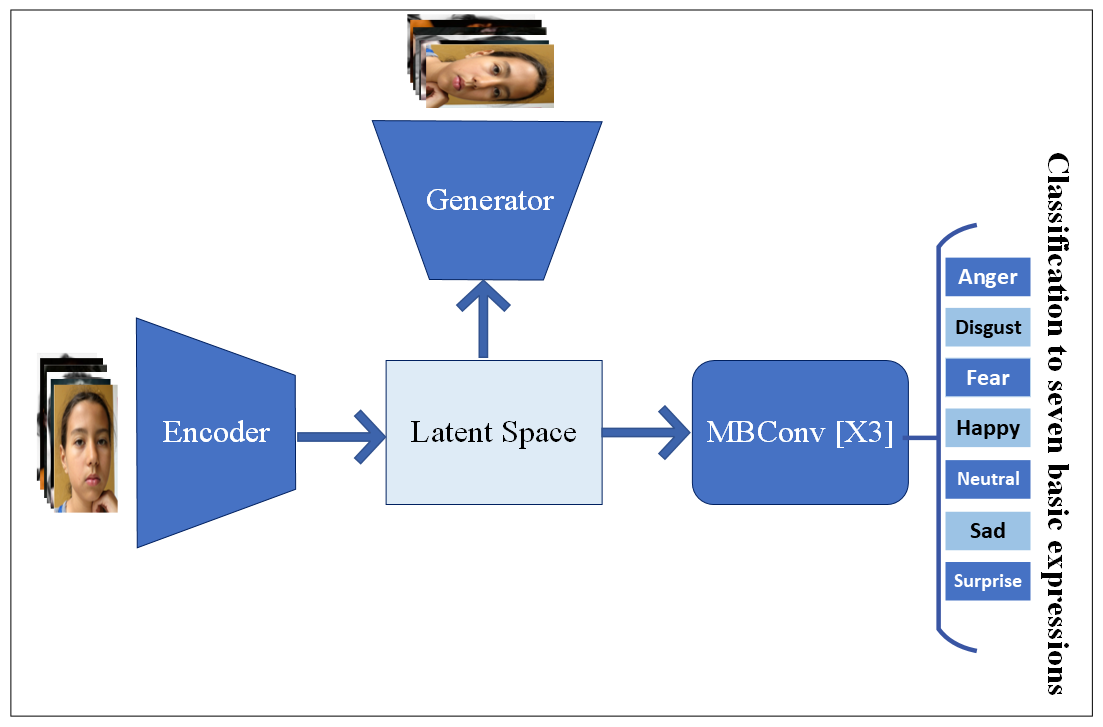}
\centering \caption{\label{fig2} Classification backbone uses the latent representation generated by the encoder to classify into the 7 emotions.}
\end{center}
\end{figure*}

We feed the latent representation generated by E into a custom classification module using MBConv\cite{mobilenet} blocks. This is demonstrated in Fig. \ref{fig2}.
\begin{equation}
\label{objective_final}
\min _{E_{\mathcal{X}_i}, G_{ \mathcal{X}_i}} \max _{D_{\mathcal{X}_i}} =\mathcal{L}_{\mathrm{VAE}}(x)+\mathcal{L}_{\mathrm{VAE}, \mathrm{D}}^{\text {latent }}(x_{q_{i}})
\quad\forall q
\end{equation}

\subsubsection{Training Method} The Encoder and the Discriminator are trained jointly with a min-max objective 
function (Equation \ref{objective_final}) with a categorical cross-entropy loss for the Discriminator.
The classification model is trained after the VAE with a symmetric cross-entropy loss for robustness.\\
\subsubsection{Training Configuration}
The training was conducted on 2 NVIDIA Tesla V100s with 32 GB of GPU memory. A Stochastic Gradient Descent Optimizer with a learning rate set to 0.0001 and momentum set to 0.9 was used. Hyper-parameter $\alpha$ from $L_{VAE}$ from Equation 1 in the paper was set to 10 after grid search.\\\\
RAF-DB \cite{r3} provides images resized to 128x128 pixels. We applied basic augmentations to our dataset, including horizontal flips with a probability of 50\% and random rotations by a maximum angle of 15\textdegree.\\
\subsubsection{Loss Functions }The proposed model has a novel loss function (Equation\ref{VAE_Objective}), which consists of three parts.
The first part is the KL Divergence between the latent and a sample from a Gaussian distribution with mean 0 and variance 1 according to \cite{r9}. This is used to provide denser representations in the latent space, improving accuracy and mitigating bias (as shown later in Section \ref{ablation}).\\
The second component is the loss from the discriminator's ability to predict the protected attribute accurately. The Encoder's goal is to be able to fool the discriminator into not knowing the protected attribute. This is the main component that aligns the latent spaces and ensures the Encoder does not learn the protected attribute features.\\
The final component is the Style-Reconstruction Loss from \cite{r5}, which is added to ensure that the semantic emotion-level features are not lost on the Generator's reconstruction of the image. This is used instead of a pixel-wise loss because expression is a subjective concept, and a pixel-wise loss does not necessarily represent semantic consistency.
\begin{equation}
\label{Perceptual_Component}
G_j^\phi(x)_{c, c^{\prime}}=\frac{1}{C_j H_j W_j} \sum_{h=1}^{H_j} \sum_{w=1}^{W_j} \phi_j(x)_{h, w, c} \phi_j(x)_{h, w, c^{\prime}}
\end{equation}
Equation \ref{Perceptual_Component} is the Gram matrix of the $j_{th}$ feature map for a network $\phi$ where $\phi_j(x)$ represents the activations of the $j_th$ layer of the network. The final loss is the squared Frobenius norm of the input and output feature matrices.\\
\subsubsection{Classification Model:}We have used 3 sequential MBConv \cite{r7} modules which use the latent representation generated by the Latent Alignment VAE and classify it into the seven basic expressions. The MBConv block has been extensively explored in many areas of deep learning and is a versatile and efficient building block. We have also experimented with using Residual Blocks \cite{r8} and found that they have a minor reduction in accuracy (described further in Section \ref{ablation}).\\

\section{Experminenation, Results, and Analysis}
\label{sec:exp}
\subsection{Evaluation Metric}
We formulate our metric for fairness as \cite{r1} and use the “equal odds” philosophy.
\begin{equation}
\label{bias_eqn}
\begin{aligned}
\mathcal{F}=\min&(\frac{\sum_{c=1}^C p\left(\hat{y}=c \mid y=c, q=q_i, \mathbf{x}\right)}{\sum_{c=1}^C p(\hat{y}=c \mid y=c, q=d, \mathbf{x})}.
 \end{aligned}
\end{equation}
\begin{center} 
 $\forall i\in (1,2....N)$\end{center}
 In equation \ref{bias_eqn}, "$d$" is the protected attribute that has the highest accuracy. We add the accuracy for each class per attribute and use the minimum value as our metric for fairness. For completeness, we also use the mean per-class per-attribute accuracy as in \cite{r6}.
 \subsection{Experimentation and Analysis}
  
Experimentation was conducted on the RAF-DB  \cite{r3} and CelebA \cite{celeba} datasets similar to \cite{r1}. The RAF-DB dataset has 7 human-annotated classes. The model is trained on the provided train-test split consisting of 12271 train images, and inference is run on 3068 test images. Table \ref{tab:table1} and Table \ref{tab:table4} show that our model achieves state-of-the-art results on RAF-DB for both metrics and demonstrates significant bias mitigation.\\
Our methodology and setup is based on the hypothesis that protected attributes can be forgotten without information loss of other facial attributes. Ideally, a network would be able to perfectly distinguish attributes if these attributes were completely separable from the rest of the informative features of the image. However, since they are not, we hypothesize that if subsets of a dataset partitioned on the basis of the protected attribute are aligned or brought closer in a latent space, these attributes are considered to be forgotten.\\ To achieve this, we use a discriminator module to classify the latents into their respective protected attributes. When this discriminator cannot determine membership of a latent into a particular protected attribute subset, then fairness can be achieved since the classification would be done solely on the basis of a latent which does not contain information about the protected attribute.
\begin{figure*}[h]
\begin{center}
\label{fig:rafdb}
\includegraphics [scale=0.23]{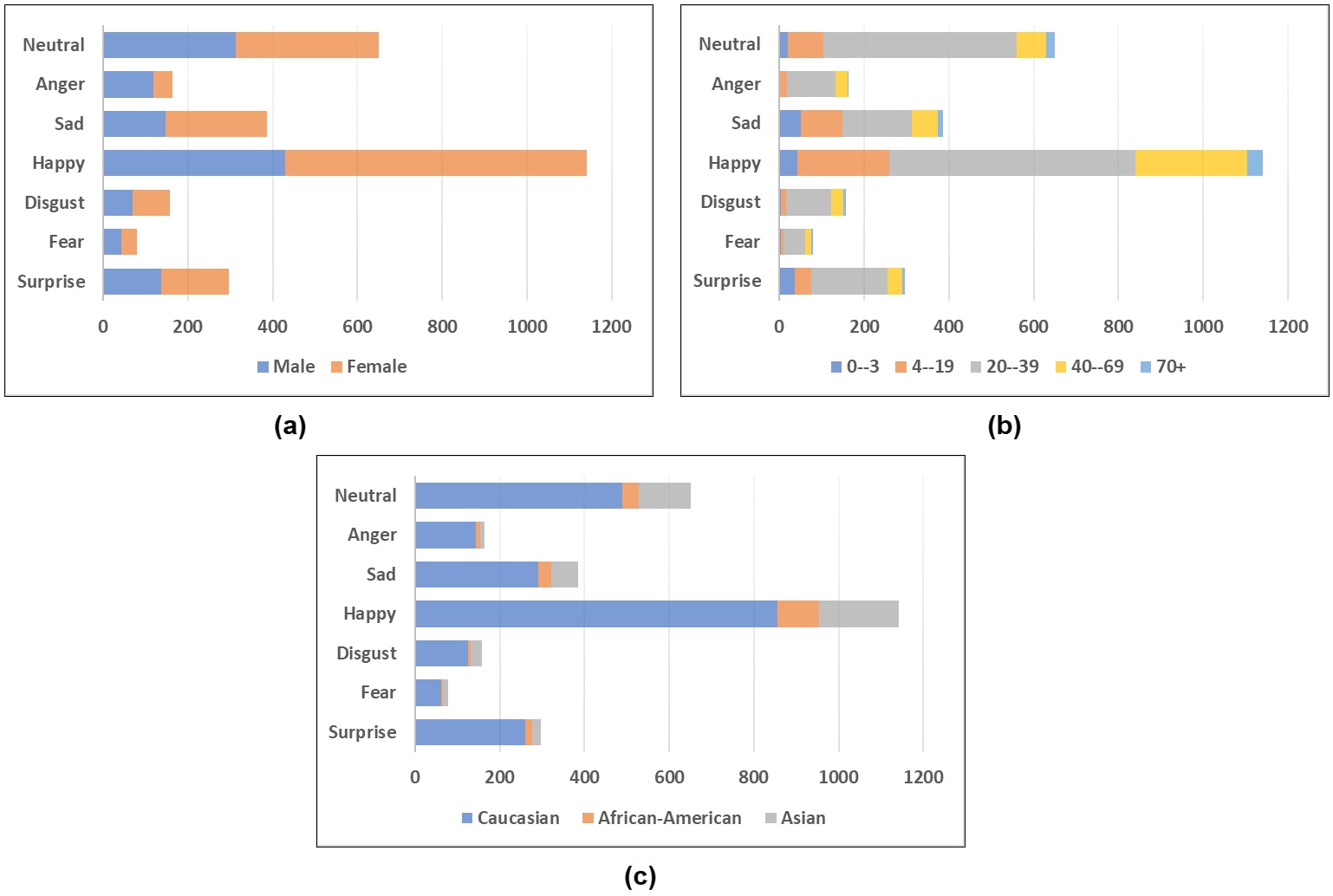}
\centering \caption{Data Distribution of the test test of RAF-DB. (a) represents the gender-wise distribution, (b) represents the age group distribution, and (c) represents the ethnic distribution of the test set of RAF-DB.}
\end{center}
\end{figure*} 
 \begin{table}[!htb]
\centering
\caption{\label{tab:table1}Comparison of expression-wise accuracies on RAF-DB.}
\begin{tabular}{lll} 
\hline
\multicolumn{1}{l|}{\multirow{2}{*}{\textbf{Expression}}} & \multicolumn{2}{c}{\textbf{Accuracy(\%)}}  \\ 
\cline{2-3}
\multicolumn{1}{l|}{}                                     & \textbf{Xu et al.} & \textbf{Ours}        \\ 
\hline
Anger~& 81.0& 83.2\\ 
\hline
Disgust& 54.1& 57.7\\ 
\hline
Fear& 53.8 & 60.2\\ 
\hline
Happy & 93.3 & 92.0\\ 
\hline
Neutral & 82.1 & 81.0 \\ 
\hline
Sad & 77.7 & 76.0 \\ 
\hline
Surprise & 81.8 & 82.9 \\ 
\hline\hline
\textbf{Mean} & \textbf{74.8} & \textbf{76.1} \\
\hline\hline
\end{tabular}
\end{table}

\begin{table}
\centering
\caption{\label{tab:table2} Mean class-wise accuracy broken down by Gender and Race attributes on RAF-DB.}
\label{tab:table2}
\resizebox{\textwidth}{!}{
\begin{tabular}{cccccccc} 
\hline
\multirow{2}{*}{\textbf{Attribute Labels}} & \multicolumn{7}{c}{\textbf{Mean Class wise Accuracies}} \\ 
\cline{2-8}
 & \textbf{Xu et al. } & \textbf{Offline\cite{m1}} & \textbf{Focal Loss\cite{m2}} & \textbf{~ DDC\cite{m3}~~} & \textbf{~ DIC\cite{m4}~~} & \textbf{~ SS\cite{ss1}~~} & \textbf{~Ours~} \\ 
\hline
Male~ & 74.2 & 72.0 & 71.0 & 71.0 & 72.0 & 72.0 & \textbf{76.3} \\
Female & 74.4 & 75.0 & 75.0 & 74.0 & 75.0 & 76.0 & \textbf{76.0} \\ 
\hline
Caucasian & 75.6 & 74.0 & 73.0 & 72.0 & 74.0 & 74.0 & \textbf{76.15} \\
African-American & 76.6 & 76.0 & 75.0 & 73.0 & 76.0 & 75.0 & \textbf{77.1} \\
Asian & 70.4 & 76.0 & 75.0 & 74.0 & 77.0 & 76.0 & \textbf{75.5} \\ 
\hline
\end{tabular}}
\end{table}

\begin{table}
\centering
\caption{\label{tab:table12} Mean class-wise accuracy broken down by Age and Gender-Race attributes on RAF-DB. }
\label{tab:table12}
\begin{tabular}{ccc} 
\hline
\multirow{2}{*}{\textbf{Attribute Labels}} & \multicolumn{2}{c}{\textbf{Mean Class wise Accuracies}} \\ 
\cline{2-3}
 & \textbf{Xu et al. } & \textbf{~Ours~} \\ 
\hline
0-3 & 80.2 & \textbf{82.4} \\
4-19 & 69.9 & \textbf{72.3} \\
20-39 & 76.4 & \textbf{77.0} \\
40-69 & 74.4 & \textbf{75.7} \\
70+ & 62.2 & \textbf{70.1} \\ 
\hline
M-Caucasian & 74.5 & \textbf{76.0} \\
M-African-American & 80.2 & \textbf{81.1} \\
M-Asian & 70.2 & \textbf{73.4} \\
F-Caucasian & 75.5 & \textbf{76.2} \\
F-African-American & 87.6 & \textbf{81.1} \\
F-Asian & 69.0 & \textbf{71.7} \\
\hline
\end{tabular}
\end{table}

\subsubsection{RAF-DB Bias Analysis.}
Most FER datasets do not have the respective age, gender, and ethnic labels; therefore, to substantiate our results, we conducted experiments on RAF-DB \cite{r3}, one of the most popular benchmark FER datasets.\\
RAF-DB contains 15,339  images of diverse facial expressions downloaded from the internet and annotated manually by crowd-sourcing and reliable estimation; this dataset consists of seven basic expressions and eleven compound expressions.\\

RAF-DB provides labels that include expression, gender type, ethnicity, and age group. Fig. 3 showcases the attribute-wise breakdown of each label class in the test data.
Since the distribution of test and training data is kept similar, we can draw few inferences from this distribution.\\
\begin{itemize}
    \item Considering "race" as an attribute, we observe that almost 77\% of the images belong to a single class i.e. Caucasian, rest, 23\% are then distributed among two attributes, namely African-American and Asian. 
    
    \item Similarly, for the age attribute, almost 57\% of the images belong to one of the five age brackets, namely \{20-39\}. The rest of the 43\% of images are distributed among the remaining four classes. Moreover, senior citizens from the 70+ age bracket and infants from \{0-3\} age bracket are highly under-represented, consisting of about 3\% and 5\%  of the total images, respectively.

    \item Observing the expression attribute, we can infer that 39.7\% of the total images belong to one of the seven expression classes, i.e. happy; the rest of the six classes are then distributed among the remaining six expressions. Moreover, expressions like fear, disgust and surprise are highly under-represented, consisting of about 2.7\%, 5\% and 10\% of the total images, respectively. 
\end{itemize}
\begin{table}
\centering
\caption{ \label{tab:table3} Comparison of mitigation of bias (higher is better) on RAF-DB broken down by attribute labels.}
\label{tab:table3}
\resizebox{\textwidth}{!}{
\begin{tabular}{cccccccc} 
\hline
\multirow{2}{*}{\textbf{Protected attributes }} & \multicolumn{7}{c}{\textbf{Mitigation of Bias}} \\ 
\cline{2-8}
 & \textbf{Xu et al.\cite{r1}} & \textbf{Offline\cite{m1}} & \textbf{Focal Loss\cite{m2}} & \textbf{DDC\cite{m3}~} & \textbf{~DIC\cite{m4}} & \textbf{SS\cite{ss1}} & \textbf{Ours} \\ 
\hline
Gender & \textbf{99.97} & 95.4 & 96.1 & 96.2 & 95.4 & 95.4 & 99.51 \\ 
\hline
Race & 91.9 & 97.4 & 97.2 & \textbf{97.6} & 96.5 & 97.5 & 94.2 \\ 
\hline
Age & 82.1 & - & - & - & - & - & \textbf{84.8} \\
\hline
\end{tabular}}
\end{table}
This further substantiates our claim and establishes the need to mitigate bias in most FER datasets. The expression accuracy shown in Table \ref{tab:table1} does not sufficiently portray the performance variation of classifiers across different demographics; therefore, in Table \ref{tab:table2},\ref{tab:table12}, we comprehensively compare accuracies broken down by each demographic group. Furthermore, to substantiate the inter-play of "gender" and "race" attributes we also provide results of joint "Gender-Race" groups in Table \ref{tab:table12}. From Table \ref{tab:table2},\ref{tab:table12} it can be inferred, that our proposed method outperforms for mean class-wise accuracies broken down by attributes "age", "gender", "race" and "gender-race".To provide a numerical assessment of mitigation of bias for sensitive attributes such as age, gender, and race, in Table \ref{tab:table3}, we provide comparisons with \cite{r1,m1,m2,m3,m4,ss1} using our evaluation metric for fairness (using Equation \ref{bias_eqn}). From Table \ref{tab:table3} we can infer that with regards to bias mitigation, our approach performs almost at par with Xu et al. \cite{r1} for "gender" attribute, whereas for "age" class it outperforms \cite{r1}.

\begin{table}[h]
\centering
\caption{\label{tab:table4}Comparison of accuracy broken down by smiling attribute on CelebA dataset. }
\begin{tabular}{ccc} 
\hline
\multicolumn{1}{c|}{\multirow{2}{*}{Expression}} & \multicolumn{2}{c}{Accuracy} \\ 
\cline{2-3}
\multicolumn{1}{c|}{} & Xu et al. \cite{r1} & Ours \\ 
\hline
Smiling~ & 92.2 & 92.9 \\
Not-Smiling & 94.1 & 94.8 \\ 
\hline
Mean & 93.15 & 93.85 \\
\hline
\end{tabular}
\end{table}
\begin{table}[h]
\centering
\caption{\label{tab:table5}Mean class-wise accuracy broken down by attributes on CelebA. }
\begin{tabular}{ccc} 
\hline
\multirow{2}{*}{Attribute Labels} & \multicolumn{2}{c}{Mean Class-wise Accuracy} \\ 
\cline{2-3}
 & Xu et al.\cite{r1} & Ours \\ 
\hline
Female~ & 93.6 & 94.5 \\
Male & 91.9 & 93.4 \\ 
\hline
Old~ & 91.6 & 92.5 \\
Young & 93.6 & 94.3 \\ 
\hline
Female-Old & 92.7 & 93.3 \\
Female-Young & 93.8 & 94.9 \\
Male-Old & 90.7 & 92.1 \\
Male-Young & 92.8 & 93.7 \\
\hline
\end{tabular}
\end{table}
\subsubsection{CelebA Bias Analysis}\hfill\\
We conduct experimentation for images in CelebA for the binary attribute of "smiling". This is done to facilitate the expression recognition of happy. We use the CelebA dataset since it is much larger as compared to RAF-DB with 39920 images in a subset of CelebA as compared to 12271 in all of RAF-DB. 
The protected attributes we use for fairness are Gender and Age.\\
The Smiling/No Smiling attribute is evenly distributed with exactly 50\% of the images having the smiling attribute. The image distribution for Gender and Age are not evenly distributed, with a 22.8\% gap between the number of Male and Female images, and a 51.4\% gap between the number of Young and Old images.
The comparison of accuracies with "Smiling" vs "No Smiling" is shown in Table \ref{tab:table4}. Since this is a simple binary classification task, accuracies are almost comparable. Table \ref{tab:table5} provides comparable class-wise (i.e. "Smiling" vs "No Smiling") accuracies broken down by attribute labels ("gender", "age", and "Gender-Age"). Table \ref{tab:table6} provides comparisons with \cite{r1} using our evaluation metric for fairness (using Equation \ref{bias_eqn}) on sensitive attributes.
\begin{table}[h]
\centering
\caption{\label{tab:table6}Comparison of mitigation of bias (higher is better) on CelebA broken down by attribute labels. }
\begin{tabular}{ccc} 
\hline
\multirow{2}{*}{Protected Attribute} & \multicolumn{2}{c}{Mitigation of Bias} \\ 
\cline{2-3}
 & Xu et al.\cite{r1} & Ours \\ 
\hline
Gender & 98.3 & 99.1 \\ 
\hline
Age & 98.1 & 98.9 \\ 
\hline
Gender-Age & 96.9 & 98.0 \\
\hline
\end{tabular}
\end{table}
\section{Ablation Study}
\label{sec:ablation}
 \label{ablation}

\begin{table}[h]
\centering
\caption{\label{abln}Component-wise Ablation Study of our model.}
\resizebox{\textwidth}{!}{
\begin{tblr}{
  cells = {c},
  hlines,
}
\textbf{Component} & \textbf{Mean Accuracy} & \textbf{Bias (Gender)} & \textbf{Bias (Race)}\\
\textbf{VAE+MBConv+Discriminator (Ours)} & 76.1 & 99.93 & 94.2\\
Auto Encoder+Discriminator+MBConv & 74.2 & 97.6 & 91.2\\
VAE+Discriminator+ResBlock & 74.5 & 99.91 & 93.8\\
VAE+MBConv & 76 & 91.4 & 79.2\\
VAE+ResBlock & 73 & 91 & 79.3
\end{tblr} }
\end{table}

We demonstrate the importance and effectiveness of each technical contribution through this ablation study on RAF-DB \cite{r3}. We first look at the impact of using a Variational Autoencoder as compared to a standard
Autoencoder or other dimensional reduction techniques. We can see a significant drop in accuracy and a corresponding drop in bias mitigation when an Autoencoder is used in place of a VAE. We believe this is due to the ability of VAEs to generate denser representations due to the KL-Divergence loss from the Gaussian distribution present in VAEs.\\

We further look at the impact of the Discriminator module on latent space alignment and examine whether it increases fairness. From Table \ref{abln}, we see that there is a significant decrease in bias mitigation when the VAE is trained without the min-max objective jointly with the discriminator. This demonstrates that the Discriminator is highly impactful for latent space alignment and that the sensitive attributes are encoded in the latent without it.\\
We further analyze the impact of the CNN classifier backbone on accuracies. We find that the MBConv block\cite{mobilenet} performs superior as compared to ResBlock \cite{r8}. In recent works, MBConv blocks have been known for their superior expressive power in CNNs. MBConv outperforms ResBlocks given all other parameters remain the same.
However, this difference is minimal given that the largest contributor to our model is the VAE+Discriminator architecture for latent alignment.

\section{Conclusion}
\label{sec:conc}
With the exponential increase of real-world artificial intelligence systems deployed in our daily lives, accounting for fairness has become a crucial factor in the design and research of such systems. AI systems can be deployed in various critical settings to make important life-changing decisions; hence, ensuring that these decisions do not exhibit bias or discriminatory behaviour against specific groups or demographics is of utmost importance. As a result, bias mitigation investigation and its mitigating strategies have gained good traction among researchers. Recently, many works have proposed bias mitigation strategies through traditional machine learning and deep learning in various subdomains; however, this is a relatively less-explored area in facial expression recognition. In this work, we propose a new method for mitigating bias in FER systems by using a Variational Autoencoder with an Adversarial Discriminator followed by an MBConv-based classification module. We surpass the results presented \cite{r1} and provide an adaptable framework that can be extended to other image classification tasks. To the best of our knowledge, this is the first work that uses latent alignment for de-biasing in FER systems. We hope that our work will pave the way for a more extensive exploration of latent space manipulation for bias reduction in a wider range of image classification scenarios.

%
%
%
\bibliographystyle{splncs04}
\bibliography{ref}

\end{document}